\documentclass[a4paper, 10pt, conference]{ieeeconf}      
\usepackage{graphicx}
\IEEEoverridecommandlockouts                              

\usepackage{amsmath,bm}
\usepackage{amssymb}
\usepackage{graphics}
\usepackage[hidelinks]{hyperref}
\usepackage{xcolor}
\usepackage{dblfloatfix}
\usepackage {tikz}
\usepackage{mathtools}
\usepackage{svg}
\usepackage{comment}
\usepackage[bottom]{footmisc}
\usepackage{tabularray}
\usepackage{amsmath}
\usepackage{soul}
\usepackage[switch]{lineno}
\usepackage{multirow}
\usepackage{caption}
\usepackage[noend]{algpseudocode}
\usepackage[linesnumbered,ruled,vlined, noend]{algorithm2e}
\usetikzlibrary {positioning}

\title{\LARGE \bf
Investigating Adaptive Tuning of Assistive Exoskeletons using \\Offline Reinforcement Learning: Challenges and Insights}

\author{Yasin Findik, Christopher Coco, and  Reza Azadeh
\thanks{Persistent Autonomy and Robot Learning (PeARL) Lab, University of Massachusetts Lowell, Lowell, MA 01854, USA. {\tt\small \{yasin\_findik, christopher\_coco\}@student.uml.edu, reza\_azadeh@uml.edu}}%
}

\begin{document}

\maketitle
\thispagestyle{empty}
\pagestyle{empty}

\begin{abstract}
Assistive exoskeletons have shown great potential in enhancing mobility for individuals 
with motor impairments, yet their effectiveness relies on precise parameter tuning for personalized assistance. In this study, we investigate the potential of offline reinforcement learning for optimizing effort thresholds in upper-limb assistive exoskeletons, aiming to reduce reliance on manual calibration. Specifically, we frame the problem as a multi-agent system where separate agents optimize biceps and triceps effort thresholds, enabling an adaptive and data-driven approach to exoskeleton control. Mixed Q-Functionals (MQF) is employed to efficiently handle continuous action spaces while leveraging pre-collected data, thereby mitigating the risks associated with real-time exploration. Experiments were conducted using the MyoPro 2 exoskeleton across two distinct tasks involving horizontal and vertical arm movements. 
Our results indicate that the proposed approach can dynamically adjust threshold values based on learned patterns, potentially improving user interaction and control, though performance evaluation remains challenging due to dataset limitations.

\end{abstract}


\section{Introduction}
Assistive robotics, particularly powered exoskeletons, have emerged as a promising technology for enhancing human mobility, whether by helping individuals with disabilities, supporting the elderly in daily activities, or improving physical performance in demanding tasks~\cite{esquenazi2017powered, baud2021review, martinez2021wearable}. Effective control in these systems depends on the ability to interpret user intentions and adapt to user learning and changes in physical conditions (e.g., fatigue)~\cite{poggensee2021adaptation}. However, most exoskeleton controllers are pre-configured and remain static after initial calibration, limiting their responsiveness over time~\cite{gopura2016developments, de2023control}. While low-level controllers, such as PID controllers, effectively regulate movement, their high-level parameters often require manual tuning to align with natural human movement patterns~\cite{nasiri2021adaptive}.

This tuning process often requires expert intervention, involving repeated testing, fine-tuning, and validation to ensure the exoskeleton responds naturally to the user's movements~\cite{pilarski2012dynamic}. Since user needs can change over time due to factors like fatigue, motor learning, or varying environments, frequent recalibration may be necessary, making the process time-consuming, resource-intensive, and difficult to implement on a larger scale~\cite{slade2022personalizing}. Addressing these challenges requires advanced modeling approaches capable of capturing the dynamic interplay between users and robotic systems. Unlike traditional rigid-body models, assistive robots function in complex, human-in-the-loop environments, requiring data-driven techniques for more adaptive and responsive control~\cite{gopinath2016human}. Consequently, research has increasingly focused on developing automated hyper-parameter tuning methods to enhance the adaptability and usability of these systems~\cite{coco2024design}.

\begin{figure}[t]
\centering
\includegraphics[width=.90\linewidth]{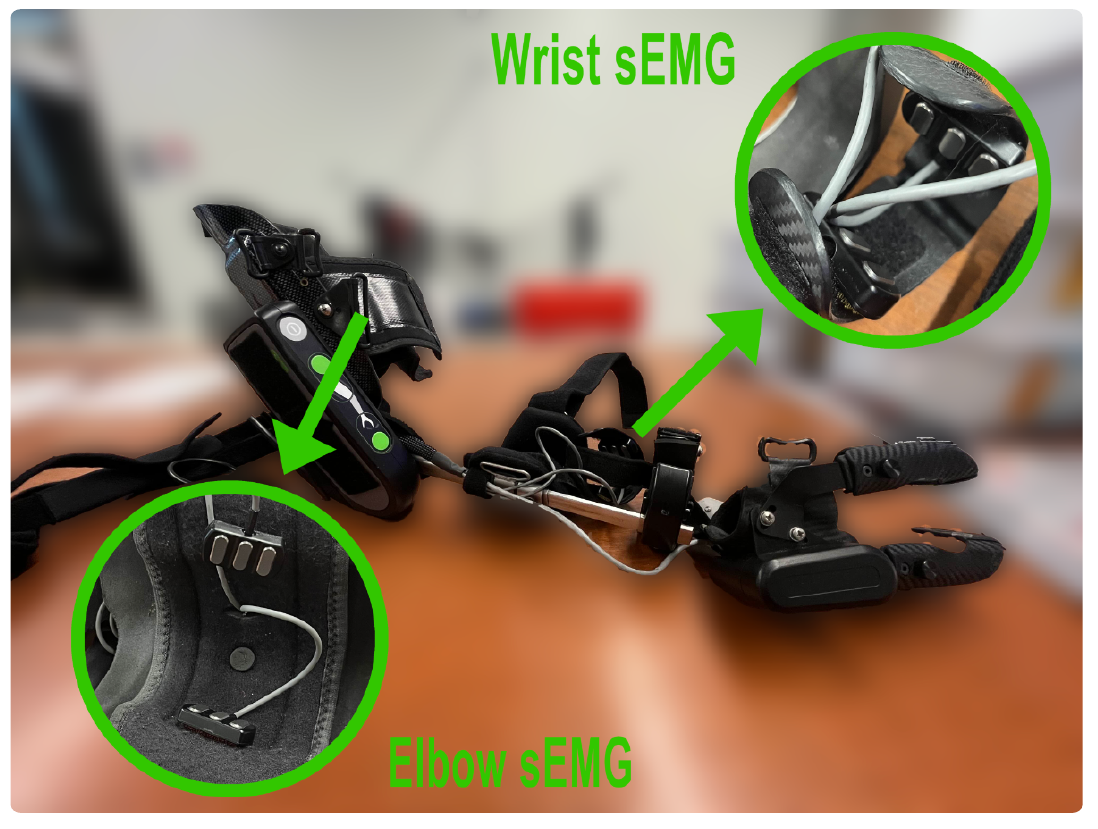}
  \caption{\small{Overview of the MyoPro 2 device, highlighting the placement of sEMG sensors.}} 
  \label{myopro}
\end{figure}

Building on these advancements, we explore a reinforcement learning-based approach to automate parameter tuning and reduce the reliance on manual recalibration. To investigate this, we use the MyoPro 2~\cite{myomo} device, a 2-DoF exoskeleton, as our experimental platform. Designed to restore functionality in individuals with paralyzed or weakened upper limbs, this device assists elbow and hand movements, as shown in Fig.~\ref{myopro}. To enhance user experience, it is essential to properly tune the robot’s hyper-parameters, particularly the biceps and triceps effort thresholds, in a way that adapts to individual users and task variations. We approach this problem through a multi-agent framework, where each agent is responsible for adjusting a specific threshold across two tasks. Specifically, we investigate the potential of offline multi-agent reinforcement learning that leverages Mixed Q-Functionals (MQF)~\cite{findik2024mixed}, enabling automated, task-specific parameter adaptation without requiring manual intervention. This approach aims to enhance the system’s adaptability and responsiveness, ultimately improving user experience and control precision.

\section{Exoskeleton Overview}

\subsection{Device Description}
The MyoPro 2 (Fig.~\ref{myopro}) is a motorized upper-limb exoskeleton designed to assist users in performing daily activities. It features two motor-controlled degrees of freedom (DoF): the elbow joint, responsible for flexion and extension, and the wrist joint, enabling hand opening and closing. The device translates muscle activation patterns into motor commands via two surface electromyography (sEMG) sensors. It is important to emphasize that this study focuses exclusively on continuous arm movements controlled by the elbow joint.

\subsection{Control Modes}
The MyoPro 2 operates in four distinct control modes:
\begin{itemize}
    \item \textit{Standby Mode:} Both motors remain inactive, ignoring sEMG sensors inputs.
    \item \textit{Biceps Mode:} The device responds solely to the biceps sEMG sensor, allowing only flexion movement.
    \item \textit{Triceps Mode:} The device responds only to the triceps sEMG sensor, allowing only extension movement.
    \item \textit{Dual Mode:} The device receives input from both sensors, enabling both flexion and extension movements.
\end{itemize}

\begin{figure}[t]
\centering
\includegraphics[width=\linewidth]{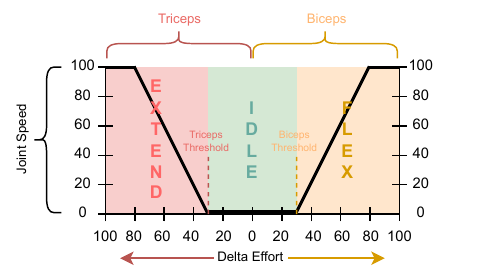}
  \caption{\small{Illustration of joint speed as a function of delta effort, highlighting the relationship between biceps and triceps activation. The graph is divided into three regions: \textit{flex}, where triceps activation exceeds the threshold, \textit{extend}, where biceps activation surpasses the threshold, and \textit{idle}, where neither muscle reaches activation. The placement of triceps and biceps thresholds is marked, demonstrating their influence on movement transitions.}} 
  \label{joint_speed}
\end{figure}

\begin{figure}[b]
\centering
\includegraphics[width=\linewidth]{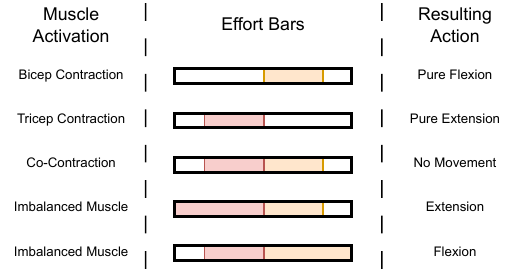}
  \caption{\small{Visualization of muscle activation patterns, corresponding effort levels, and resulting actions. The left column categorizes different activation scenarios, including biceps contraction, triceps contraction, co-contraction, and imbalanced muscle activation. The middle column represents effort levels using effort bars, where biceps and triceps contributions are indicated. The right column shows the resulting action, such as pure flexion, pure extension, no movement, or imbalanced extensions and flexions, based on the activation pattern.}} 
  \label{effort_bars}
\end{figure}

This study focuses on the \emph{Dual Mode}, as it aligns with the natural bidirectional control required for most daily tasks. Within this mode, three movement types can be selected: constant, proportional, and ramp (i.e., exponential). We specifically examine proportional mode, where movement speed scales proportionally to the user's muscle effort.

In proportional mode, the joint speed, $S_\textrm{joint}$, is determined by the difference in efforts between the dominant and opposing muscles:
\begin{align}
    S_\textrm{joint} = k_p  \Delta E,     \nonumber
\end{align}
\noindent where $k_p$ is proportional gain value specified by the manufacturer, and $\Delta E$ is the difference between the dominant and opposing muscle efforts, $E_d$ and  $E_o$,  respectively. The muscle exhibiting the higher activation level is considered dominant. Once the delta effort, $\Delta E= E_d - E_o$, exceeds the predefined threshold for the dominant muscle, $S_\textrm{joint}$ increases proportionally to $\Delta E$ in the corresponding direction, up to a maximum speed of $100$ (dimensionless), as illustrated in Fig~\ref{joint_speed}. For example, if a user intends to flex their arm, the biceps effort ($E_d$) excees the triceps effort ($E_o$), generating a positive $\Delta E$. If this difference surpasses the threshold, the exoskeleton initiates flexion at a speed proportional to $\Delta E$. Conversely, if the triceps effort becomes dominant, the device facilitates extension. The possible muscle activations and their corresponding actions are illustrated in Fig.~\ref{effort_bars}.

\subsection{Effort Threshold Optimization}

To summarize, beyond selecting a movement type, users can adjust two key parameters: (a) the effort thresholds, which define the biceps/triceps activation levels required to initiate movement, and (b) the gain values, which amplify the sEMG signal. In this study, we focus on optimizing effort thresholds rather than gain values. While gain values affect sensor sensitivity, effort thresholds play a more direct role in determining when and how the device responds to muscle activation. Properly tuning these thresholds enhances control precision, ensuring smoother and more intuitive user-device interaction.

\section{Proposed Method}

Offline Reinforcement Learning (RL)~\cite{levine2020offline} is a powerful data-driven approach for optimizing sequential decision-making. Unlike traditional Deep Reinforcement Learning (DRL), where an agent interacts with the environment to collect experience, offline RL relies solely on a fixed dataset collected through an arbitrary process, eliminating the need for online exploration~\cite{lange2012batch}. This characteristic makes offline RL particularly well-suited for scenarios where real-time interaction is impractical, costly, or unsafe. In the context of exoskeleton control, where continuous online experimentation could lead to user discomfort or even injury, offline RL provides a viable alternative by leveraging pre-collected data to optimize control policies. This enables adaptive and personalized assistance while minimizing risks associated with direct user interaction during training.

The applications of exoskeletons vary depending on the user's condition, often necessitating complex control strategies that could benefit from multi-agent systems. Although our study focuses on a single upper-limb exoskeleton, we frame the problem as a multi-agent system by assigning one agent to optimize the biceps effort threshold and another to optimize the triceps effort threshold. This formulation not only allows for more flexible and adaptive control but also provides a scalable framework that can be extended to scenarios where multiple exoskeletons or assistive devices operate collaboratively to enhance user mobility.

\begin{figure}[t]
\centering
\includegraphics[width=\linewidth]{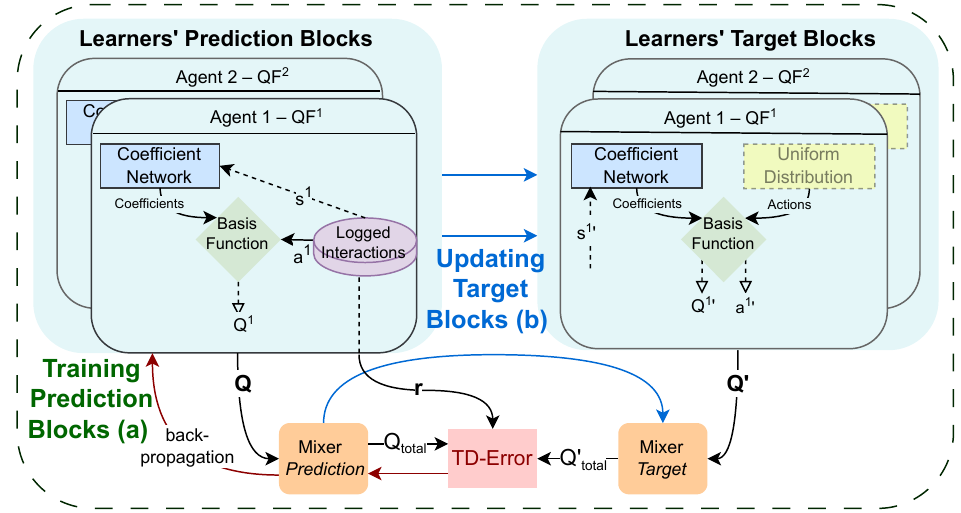}
  \caption{\small{Overview of the proposed offline MARL architecture with MQF: (a) training prediction blocks, (b) updating target blocks. The red arrows indicate the direction of the backpropagation, while the blue arrows depict the target network updates for mixer and learners' prediction blocks.}} 
  \label{overview}
\end{figure}

Since the threshold values serve as the action space for the agents and take continuous values, specifically ranging between $[20, 50]$ for the MyoPro device, we employ Mixed Q-Functionals (MQF)~\cite{findik2024mixed} as the learning algorithm for our multi-agent system. MQF has demonstrated superior performance compared to policy-based algorithms, particularly in continuous settings, due to its increased sample efficiency, an essential factor in offline learning where data collection is limited and costly. By leveraging MQF, our approach ensures more effective learning from pre-collected data, leading to improved threshold optimization for adaptive exoskeleton control.

MQF transforms each state into a set of parameters that define a function over the action space. This approach represents states through learned coefficients of basis functions, enabling efficient evaluation of multiple actions via matrix operations. A key advantage of MQF over other value-based multi-agent reinforcement learning methods is its ability to handle continuous action spaces while maintaining the sample efficiency inherent in value-based approaches, resulting in improved agent performance compared to policy-based methods. In this work, we adapt MQF for offline settings by modifying its structure. As illustrated in Fig.~\ref{overview}, our proposed framework consists of two main components: (i) training learners' prediction blocks, and (ii) updating learners' target blocks.

\noindent\textbf{(i) Training Learners' Prediction Blocks:}
The training process for the MQF begins with the selection of a batch of transition tuples $\langle s^i, a^i, r^i, {s'}^i\rangle$, each of size $b$, from the logged interactions $L^i$ of each agent. These tuples are utilized to calculate the agents' individual $Q$ and ${Q'}$ values. Concisely, the process starts with the agent's coefficient neural network predicting the basis function coefficients for the current state. In simple terms, each state is mapped to a function within the action domain, with the function’s coefficients predicted by the network, $C(s)$. These coefficients are then used to calculate action values by multiplying them with the corresponding action representations. The total number of coefficients, $k$, is determined using the combinatorial formula ${\mathcal{O} + D \choose D}$, where $D$ represents the action dimension and $\mathcal{O}$ the order of the state function. Thus, the $Q$ function can be expressed as follows:
\begin{align*}
Q^\textrm{F}(s, a) = C(s)^\top \Phi(a),
\end{align*}
where $C(s)$ returns a vector of coefficients, calculated by the neural network and $\Phi(a)$ returns the action representations as a vector, computed according to the predefined function type.

For simplicity, assume the basis function is a \textit{polynomial} of order $2$, and the action space is $2$-dimensional $a=[a_1, a_2]$. Under these conditions, the returned coefficient vector can be represented as:
\begin{align*}
 C(s)=\begin{bmatrix}
        c_{0}&
        c_{1}&
        c_{2}&
        c_{3}&
        c_{4}&  
        c_{5}
 \end{bmatrix}^\top.
\end{align*}
The second-order \textit{polynomial} basis consists of all polynomials of the form $a_1^ia_2^j$ where $i+j\leq2$. Therefore, the action representation is expressed as follows:
\begin{align*} \Phi(a)= \begin{bmatrix}
           1&
        a_{1}&
        a_{1}^2&
        a_{2}&
        a_{2}^2&  
        a_{1}a_2
 \end{bmatrix}^\top.
\end{align*}
And, $Q$ value for given state-action pair becomes as follows:
\begin{align*} Q^\textrm{F}(s, a) = c_0 + c_1a_1 + c_2a_1^2 + c_3a_2 + c_4a_2^2 + c_5a_1a_2.
\end{align*}
By leveraging matrix multiplication, the evaluation of uniformly sampled actions is significantly accelerated, allowing the action with the highest $Q$-value, in~\eqref{e_td}, to be selected more efficiently.

\setcounter{figure}{5}
\begin{figure*}[b]
\centering
\includegraphics[width=0.9\linewidth]{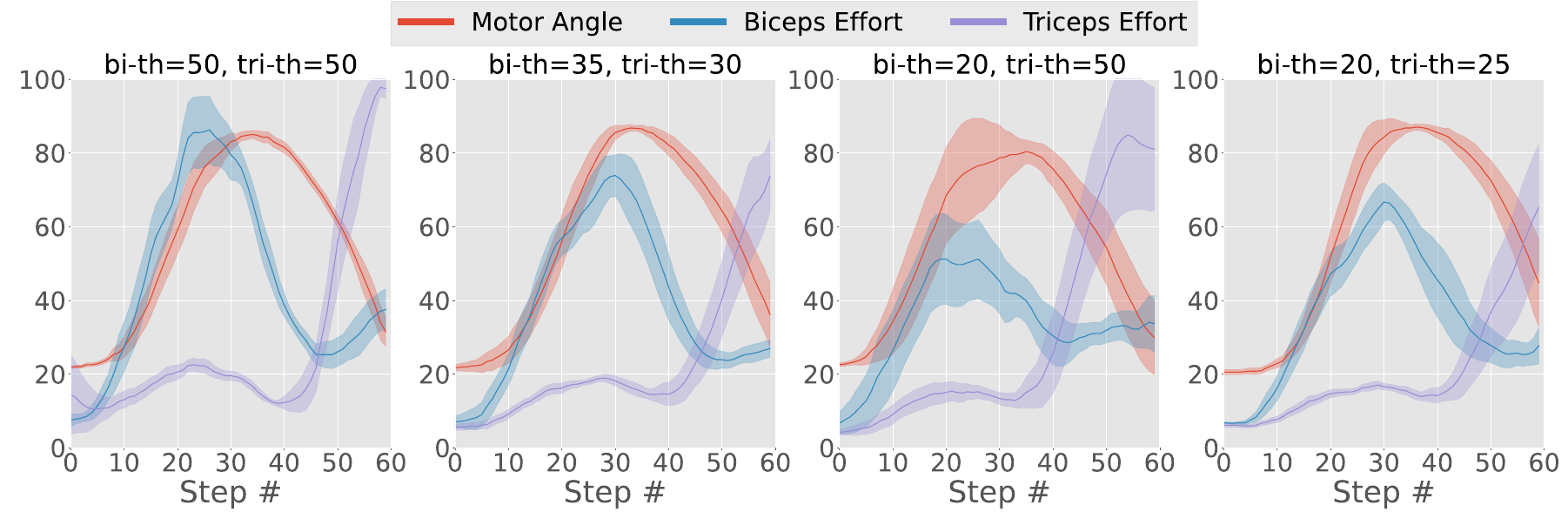}
  \caption{\small{Visualization of data collected from the vertical task (Fig.~\ref{tasks}(b)), showing the relationship between \textit{motor angle}, \textit{biceps} effort, and \textit{triceps} effort across different step numbers. Each subplot represents a different combination of biceps (bi-th) and triceps (tri-th) effort thresholds, as indicated in the titles. The red, blue and purple lines represents motor angle, biceps effort, triceps effort, respectively, with shaded regions indicating the 95\% confidence interval. As this task primarily relies on biceps activation, lower biceps threshold values (e.g., 20) appear to facilitate more efficient movement by reducing the required effort while preserving stability and control.}} 
  \label{results_vertical}
\end{figure*}

We employ another two networks: one for mixing the agents' current $Q$ values to derive $Q_{\textrm{total}}$, and another for mixing the next-state $Q'$ values to obtain $Q'_{\textrm{total}}$. Fig.~\ref{overview}(a) illustrates these networks as the mixer prediction and the mixer target, respectively. This procedure is followed by the calculation of the temporal difference error between the prediction and target values, which can be encapsulated by the following formulation:
\begin{align}
\label{e_td}
e_{\textrm{TD}} &= \sum_{\bm{s}, \bm{a}, \bm{r}, \bm{s}' }^{\bm{b} \sim \bm{B}} [(Q_{\textrm{total}}(\bm{s}, \bm{a}; \bm{w}_{\textrm{PN}}) - y(\bm{r},\bm{s}'))^2], 
\end{align}
\noindent where $y(\bm{r},\bm{s}')$ is defined as:
\begin{align}
\label{y}
y(\bm{r},\bm{s}') &=\sum_{i=1}^{N} \bm{r}^i + \gamma \max_{\bm{a}'}(Q_{\textrm{total}}(\bm{s}', \bm{a}'; \bm{w}_{\textrm{TN}})|_{\bm{a}'\sim \mathcal{U}}).
\end{align}
\noindent  
The bold symbols, in \eqref{e_td} and \eqref{y}, are vectors denoting the set of corresponding values for all agents and $\mathcal{U}$ signifies a uniform distribution. The function \smash{$Q_{\textrm{total}}$} serves as a mixing mechanism for the agents' $Q$-values. It is also important to note that we view the function for mixing agents' action values as a meta-function that can vary from additive forms, as in Value Decomposition Networks (VDN)~\cite{sunehag2017value}, to monotonic functions, as in QMIX~\cite{rashid2018qmix}, or more complex methods, similar to those described in Qtran~\cite{son2019qtran} and weighted 
QMIX~\cite{rashid2020weighted}. A more complex mixer model requires increased training effort to handle task complexities. However, for the tasks presented in this study, we observed that simpler mixer functions, as in VDN, work well. The optimization of the agents' Q-functionals aims to minimize the temporal difference error, $e_{\textrm{TD}}$ in \eqref{e_td}, thus facilitating the coordination of agent actions to maximize collective rewards. This training is conducted centrally using \smash{$Q_{\textrm{total}}$}, while the determination of individual agent actions is guided by their \smash{$Q_{\textrm{F}}^i$}.

\setcounter{figure}{4}
\begin{figure}[t]
\centering
\includegraphics[width=0.9\linewidth]{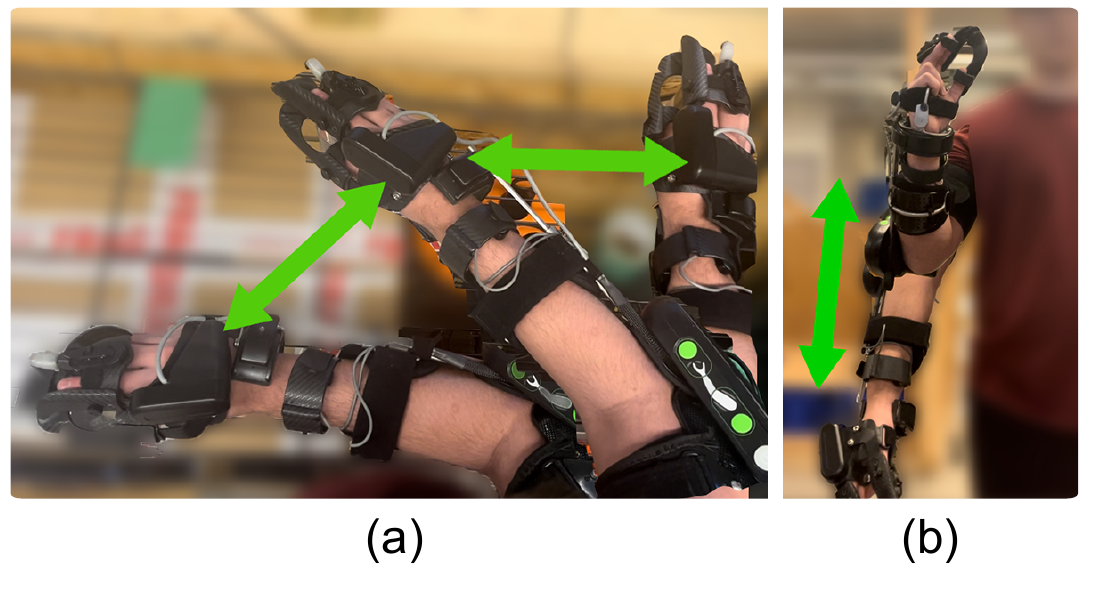}
  \caption{\small{Demonstration of task movements for data collection using the MyoPro exoskeleton. (a) Depicts the \textit{horizontal} task, where the user moves their arm laterally. (b) Shows the \textit{vertical} task, involving an upward and downward arm motion.}} 
  \label{tasks}
\end{figure}

\vspace{0.3em}
\noindent\textbf{(ii) Updating Learners' Target Blocks:}
Our framework, designed for multi-agent environments with continuous action spaces, diverges from standard value-based methods in updating the target networks. It employs a \textit{soft} update mechanism at each time-step, proven more effective in continuous action domains instead of \textit{periodic} updates~\cite{lillicrap2015continuous}. This method is characterized by incremental adjustments, formalized as:
\begin{align*}
\label{smooth_update}
w_{\textrm{TN}} = \tau w_{\textrm{PN}} + (1-\tau)w_{\textrm{TN}},
\end{align*}
\noindent 
where $\tau$ is a small factor ($\tau \ll 1$), ensuring improved stability and effectiveness in such environments.

\section{Experiments}

\setcounter{figure}{6}
\begin{figure*}[t]
\centering
\includegraphics[width=.9\linewidth]{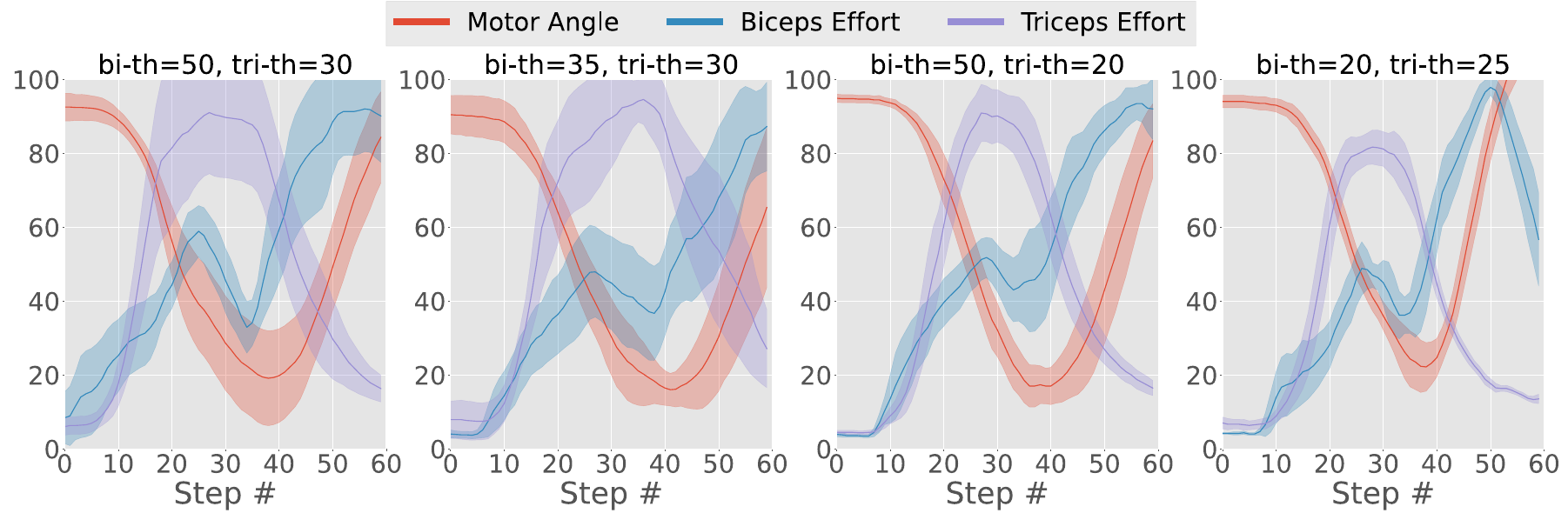}
  \caption{\small{Analysis of data collected from the horizontal task (Fig.~\ref{tasks}(a)), illustrating \textit{motor angle}, \textit{biceps} and \textit{triceps} efforts over time for different biceps (bi-th) and triceps (tri-th) effort thresholds (indicated in the titles). The red line depicts motor angle, while blue and purple lines correspond to biceps and triceps effort, respectively. Shaded regions indicate the 95\% confidence interval, highlighting variability in the recorded data. Since this task primarily relies on triceps activation, biceps and triceps threshold values $(20,25)$, respectively, appear to be the most effective, as they minimize user triceps effort while maintaining smooth movement.}} 
  \label{results_horizontal}
\end{figure*}

For our experiments, we utilize the MyoPro~\cite{myomo}, a lightweight upper-limb exoskeleton with two degrees of freedom. This wearable device is equipped with four surface electromyography (sEMG) sensors, two placed on the upper arm and two on the forearm, which detect muscle activity associated with biceps and triceps engagement. The exoskeleton's motor activates when predefined thresholds are exceeded, providing assistance accordingly. Our approach focuses on optimizing these threshold values to enhance ease of use and improve user experience. To evaluate its effectiveness, we train three models: one for each task individually and a third designed to generalize across both tasks.

\subsection{Task Design \& Data Collection}

We designed two tasks, vertical and horizontal, to capture two ubiquitous and essential primitive movement patterns. In the horizontal task, a participant transported an empty can between two fixed locations on a table, emphasizing a controlled lateral arm movement. The vertical task required rotating the arm around a shoulder-originating axis, mimicking a curling motion. These tasks were selected for their distinct characteristics: the horizontal task consists of movements along the horizontal plane, whereas the vertical task requires movements along the vertical axis, engaging different muscle groups and movement dynamics. Demonstration of both tasks are shown in Fig.~\ref{tasks}.

\subsection{Data Collection}

To collect data, each task was repeated multiple times with systematically varying the biceps and triceps effort thresholds. To limit the number of demonstrations and avoid fatigue, these thresholds were manually adjusted in increments of $5$, ranging from $20$ to $50$. For each threshold combination, data was collected over $10$ episodes, with each episode lasting approximately 40 seconds. To minimize the effects of fatigue, the user also took brief rest periods between episodes. 
Fig.~\ref{results_vertical} and Fig.~\ref{results_horizontal} illustrate the data distribution for selected thresholds in the vertical and horizontal tasks, respectively, where solid lines represent the average values, and the shaded area indicates the 95\% confidence interval.


\subsection{Results, Insights \& Challenges}

\noindent\textbf{Preprocessing -} We begin by preprocessing the collected data to ensure it is suitable for training a reinforcement learning model. The state representation consists of three variables: $\langle p,  {E_\textrm{biceps}}, {E_\textrm{triceps}} \rangle$, where $p$ represents the motor angle, and $E_\textrm{biceps}$ and $E_\textrm{triceps}$ correspond to the effort levels for the biceps and triceps, respectively. The action space consists of adjusting either the biceps threshold, $th_\textrm{biceps}$, or the triceps threshold, $th_\textrm{triceps}$, depending on the assigned agent. Specifically, the agent responsible for biceps muscle adjusts $th_\textrm{biceps}$, and the triceps agent modifies $th_\textrm{biceps}$.

\vspace{5px}
\noindent\textbf{Reward function design -} To guide learning, we define a reward function for each state-action pair, formulated as:
\begin{align*}
r = e^{\smash{-d/c}},
\end{align*}
\noindent
where $d$ represents the difference between $\Delta E$ and the threshold value of the dominant muscle, and $c$ is a constant. The objective is to minimize this difference, ensuring that the exoskeleton provides continuous and adaptive assistance throughout task execution. By incorporating the exponential function, we constrain the reward values within the range $[0,1]$, maintaining numerical stability and smooth optimization during training.

\vspace{5px}
\noindent\textbf{Training -} We have trained three models: one for the horizontal task, one for the vertical task, and one for both tasks combined. During training, batches were drawn from the preprocessed dataset, which captures how actions (i.e., threshold values) influence new states, such as pose (i.e., motor angle) and effort values. Because the dataset already contains the outcomes of specific state-action pairs, reward calculation during training requires no additional estimation. The models are trained to maximize reward values that can be directly computed using the described reward function, as all relevant parameters are known, ensuring efficient learning within the dataset’s constraints.

\addtolength{\textheight}{-5.8cm}
\vspace{5px}
\noindent\textbf{Findings - } Our results highlight the task-dependent nature of threshold optimization in assistive exoskeletons. Fig.~\ref{results_vertical} illustrates the findings from the vertical task (Fig.~\ref{tasks}(b)), where movement primarily relies on biceps activation. The data suggests that lower biceps threshold values (e.g., 20) facilitate smoother movement by reducing the required effort while maintaining control. Conversely, Fig.~\ref{results_horizontal} presents results from the horizontal task (Fig.~\ref{tasks}(b), which depends more on triceps activation. Here, biceps and triceps thresholds of $(20, 25)$, respectively, appear to be the most effective, as they minimize user effort while ensuring smooth and stable motion. These insights emphasize the importance of adaptive tuning to accommodate varying task demands. Even though the models can dynamically set thresholds for the muscles, evaluating their performance and comparing them to static threshold settings presents a significant challenge. When the models generate new actions (i.e., threshold values), they may lead to states that are not present in the dataset, making their outcomes unknown. This issue is further compounded by the dataset’s limitations---it was collected from a single participant with threshold increments of only $5$, restricting the ability to assess overall performance comprehensively. One approach to mitigate this limitation is expanding the dataset to ensure that generated actions during testing lead to states already represented in the data. Alternatively, a transition model could be trained to predict state changes resulting from new threshold values, as suggested in~\cite{osooli2024investigating}. Ultimately, the most reliable way to assess the impact of dynamic threshold adjustments on user performance in daily tasks is through real-world testing. Future work should involve multiple participants and iterative training cycles incorporating human feedback to refine the models based on real-world interactions.

\section{Conclusions \& Future Work}

This study explored the application of offline reinforcement learning for optimizing effort threshold parameters in a upper-limb assistive exoskeleton. By leveraging a multi-agent learning framework with Mixed Q-Functionals (MQF), we aimed to enable adaptive and data-driven parameter tuning, reducing the need for manual calibration. Our experimental results indicate that the proposed approach can learn from pre-collected data, enabling dynamic adjustments to effort thresholds that may improve user interaction and control. However, challenges remain in evaluating model performance due to the limitations of offline datasets and the absence of real-time interaction. Future work will focus on expanding the dataset with a more diverse participant pool, integrating transition models to better estimate unseen states, and conducting real-world user studies to assess the practical impact of dynamic threshold adjustments. Additionally, further investigations into task complexity and alternative adaptive learning techniques will be essential to improving assistive exoskeleton control in broader applications.

\section*{Acknowledgement}
This work was supported by the National Science Foundation (CMMI-2110214).

\bibliographystyle{IEEEtran}
\bibliography{IEEEabrv, refs}

\end{document}